\documentclass[11pt,letterpaper]{article}
\usepackage{emnlp2017}
\usepackage{times}
\usepackage{latexsym}

\usepackage{graphicx}
\usepackage{multirow}
\usepackage{amssymb}
\usepackage{graphicx}
\usepackage{array}
\usepackage{amsmath}
\usepackage{algorithm}
\usepackage[final]{changes}
\definechangesauthor[color=red]{lq}
\definechangesauthor[color=orange]{yl}

\usepackage{colortbl}

\usepackage{algpseudocode}
\usepackage{CJK}
\usepackage{url}

\emnlpfinalcopy

\def\emnlppaperid{439}

\title{A Question Answering Approach to Emotion Cause Extraction}

\author{Lin Gui$^{a,b}$, Jiannan Hu$^a$, Yulan He$^c$, Ruifeng Xu$^{a,d}$\footnotemark[2], Qin Lu$^e$, Jiachen Du$^a$ \\
         $^a$Shenzhen Graduate School, Harbin Institute of Technology, China
  \\    $^b$College of Mathematics and Computer Science, Fuzhou University, China   \\    $^c$School of Engineering and Applied Science, Aston University, United Kingdom    \\   $^d$Guangdong Provincial Engineering Technology Research Center for Data Science, China   \\   $^e$Department of Computing, the Hong Kong Polytechnic University, Hong Kong \\
{\tt guilin.nlp@gmail.com, hujiannan0526@gmail.com,} \\ {\tt  y.he9@aston.ac.uk, xuruifeng@hit.edu.cn, }\\{\tt csluqin@comp.polyu.edu.hk, dujiachen@stmail.hitsz.edu.cn}}

\date{}

\begin{document}

\begin{CJK*}{UTF8}{gbsn} 

\maketitle

\begin{abstract}
Emotion cause extraction aims to identify the reasons behind a certain emotion expressed in text. It is a much more difficult task compared to emotion classification. Inspired by recent advances in using deep memory networks for question answering (QA), we propose a new approach which considers emotion cause identification as a reading comprehension task in QA. Inspired by convolutional neural networks, we propose a new mechanism to store relevant context in different memory slots to model context information. Our proposed approach can extract both word level sequence features and lexical features. Performance evaluation shows that our method achieves the state-of-the-art performance on a recently released emotion cause dataset, outperforming a number of competitive baselines by at least 3.01\% in F-measure.

\end{abstract}

\section{Introduction}
\label{sec:section1}

\renewcommand{\thefootnote}{\fnsymbol{footnote}}

With the rapid
growth of social network platforms, more and more people tend to share their experiences and emotions online.\footnotetext[2]{Corresponding Author: xuruifeng@hit.edu.cn} Emotion analysis of online text becomes a new challenge in Natural Language Processing (NLP). In recent years, studies in emotion analysis largely focus on emotion classification including detection of writers' emotions~\cite{gao2013joint} as well as readers' emotions~\cite{chang2015linguistic}. There are also some information extraction tasks defined in emotion analysis~\cite{Chen2016A,Balahur2011EmotiNet}, such as extracting the feeler of an emotion ~\cite{das2010finding}. These methods assume that emotion expressions are already observed. Sometimes, \added{however,} we care more about the stimuli, or the cause of an emotion. For instance, Samsung wants to know why people love or hate Note 7 rather than the distribution of different emotions.

\noindent \textbf{Ex.1} \emph{我的手机昨天丢了，我现在很难过。}

\noindent \textbf{Ex.1} \emph{Because I lost my phone yesterday, I feel sad now.}

In an example shown above, ``sad'' is an emotion word, and the cause of ``sad'' is ``I lost my phone''. The emotion cause extraction task aims to identify the reason behind an emotion expression. It is a more difficult task compared to emotion classification since it requires a deep understanding of the text that conveys an emotions.

Existing approaches to emotion cause extraction mostly rely on methods typically used in information extraction, such as rule based template matching, sequence labeling and classification based methods. Most of them use linguistic rules or lexicon features, but do not consider the semantic information and ignore the relation between the emotion word and emotion cause. 

\begin{figure}[htbp]
\label{fig:figure1}
\centering
\includegraphics[width=3in]{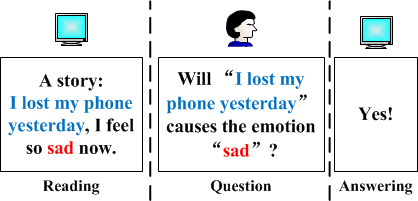}
\caption{An example of emotion cause extraction based on the QA framework.}
\end{figure}

In this paper, we present a new method for emotion cause extraction. We consider 
emotion cause extraction as a question  answering (QA) task. Given a text containing the description of an event which \added[id=lq]{may or may not} cause
a certain emotion, we take \added[id=lq]{an}
emotion word \added[id=lq]{in context}, such as ``sad'', as a query. The question to the QA system is: ``Does the described event cause the emotion of sadness?''. The \added[id=lq]{expected} answer \added[id=lq]{is}
either ``yes'' or ``no''. (see Figure \ref{fig:figure1}). We build our QA system based on a deep memory network. The memory network has two inputs: a piece of text, \added[id=lq]{referred to as a story in QA systems,} and a query. The 
\added[id=lq]{story }
is represented using a sequence of word embeddings. 

\added[id=lq]{A} recurrent structure is implemented to mine the deep relation between a query and a text. It measure\added[id=lq]{}s the \added[id=lq]{importance}
of each word in the text by \added[id=lq]{an} attention mechanism. Based on the \added[id=lq]{learned} attention result, 
the network maps the text into a low dimensional vector space. 
This vector is \added[id=lq]{then} used to generate an answer. Existing memory network based approaches to QA use weighted sum of attentions to jointly consider short text segments stored in memory. However, they do not explicitly model 
\added[id=lq]{sequential information in the context.}
In this paper, we propose a new deep memory network architecture to model the context of each word simultaneously by
multiple memory slots which capture sequential information using convolutional operations~\cite{kim2014con}, and 
achieves the state-of-the-art performance compared to existing methods which use manual rules, common sense knowledge bases or other machine learning models.


The rest of the paper is organized as follows. Section \ref{sec:section2} gives a review of related works on emotion analysis. Section \ref{sec:section3} presents our proposed deep memory network based model for emotion cause extraction. Section \ref{sec:section4} discusses evaluation results. Finally, Section \ref{sec:section5} concludes the work and outlines the future directions.

\section{Related Work}
\label{sec:section2}

Identifying emotion categories in text is one of the key tasks in NLP \cite{liu2015sentiment}. Going one step further, emotion cause extraction can reveal  important information \added{about what causes a certain emotion and why there is an emotion change}. In this section, we introduce related work on emotion \added{analysis including emotion} cause extraction. 

In emotion analysis, we first need to determine the taxonomy of emotions. Researchers have proposed a list of primary emotions \cite{plutchik1980emotion,ekman1984expression,turner2000origins}. In this study, we adopt Ekman's emotion classification scheme~\cite{ekman1984expression}, which identifies six primary emotions, namely \emph{happiness, sadness, fear, anger, disgust} and \emph{surprise}, \deleted{as} known as the ``Big6'' scheme in the W3C Emotion Markup Language. This emotion classification scheme is agreed upon by most previous works in Chinese emotion analysis.

Existing work in emotion analysis mostly focuses on emotion classification~\cite{li2013sentence,zhou2016emotion} and emotion information extraction~\cite{Balahur2013Detecting}. \newcite{xu2012coarse} used a coarse to fine method to classify emotions in Chinese blogs. \newcite{gao2013joint} proposed a joint model to co-train a polarity classifier and an emotion classifier. \newcite{beck2014joint} proposed a Multi-task Gaussian-process based method for emotion classification. \newcite{chang2015linguistic} used linguistic templates to predict reader's emotions. \newcite{das2010finding} used an unsupervised method to extract emotion feelers from Bengali blogs. There are 
other studies which focused on joint learning of sentiments~\cite{luojeam,mohtarami2013probabilistic} or emotions in tweets or blogs~\cite{quan2009construction,liu2013joint,hasegawa2013predicting,qadir2014learning,ou2014exploiting}, and emotion lexicon construction~\cite{mohammad2013crowdsourcing,yang2014topic,staiano2014depechemood}. However, the aforementioned work all focused on \added{analysis of}
emotion expressions rather than \added{emotion causes.}

\newcite{lee2010text} first proposed a task on emotion cause extraction. They \added{manually} constructed a corpus from the Academia Sinica Balanced Chinese Corpus. Based on this corpus, \newcite{chen2010emotion} proposed a rule based method to detect emotion causes based on manually define linguistic rules. Some studies~\cite{gui2014emotion,li2014text,gao2015rule} extended the rule based method to informal text in Weibo text (Chinese tweets).

Other than rule based methods, \newcite{russo2011emocause} proposed a crowdsourcing method to construct a common-sense knowledge base which is related to emotion causes. But it is challenging to extend the common-sense knowledge base automatically. \newcite{ghazi2015detecting} used Conditional Random Fields (CRFs) to extract emotion causes. However, it requires emotion cause and emotion keywords to be in the same sentence. More recently, \newcite{gui2016event} proposed a multi-kernel based method to extract emotion causes through learning from a manually annotated emotion cause dataset. 


\added[id=lq]{Most existing work does} not consider the relation between an emotion word and the cause of such an emotion, or they simply use the emotion word as a feature in their model learning. 
Since emotion cause extraction requires an understanding of a given piece of text in order to correctly identify the relation between the description of an event which causes an emotion and the expression of that emotion, it can essentially be considered as a QA task. In our work, we choose the memory network, which is designed to model the relation between a story and a query for QA systems~\cite{weston2014memory, Sukhbaatar2015end}. Apart from its application in QA, memory network has also achieved great successes in other NLP tasks, such as machine translation~\cite{Luong2015MT}, sentiment analysis~\cite{tang2016memory} or summarization~\cite{rush2015summarization}. To the best of our knowledge, this is the first work which uses memory network for emotion cause extraction. 

\section{Our Approach}
\label{sec:section3}

In this section, we will first define our task. \added[id=lq]{Then, a brief introduction of memory network will be given, including its basic learning structure of memory network and deep architecture. Last, our modified deep memory network for emotion cause extraction will be presented.} 

\subsection{Task Definition}
\label{sec:definition}

The formal definition of emotion cause extraction is given in \cite{gui2016event}. In this task, a given document, which 
\added[id=lq]{is a passage}
about an emotion event, contains an emotion word $E$ and the cause of the event. The document is manually segmented in the clause level. For each clause $c = \{w_{1}, w_{2},...w_{k}\}$ consisting of $k$ words, the goal 
\added[id=lq]{is to identify}
which clause contains the emotion cause. 
\added[id=lq]{For data representation,}
we can map each word into a low dimensional embedding space, a.k.a word vector~\cite{mikolov2013distributed}. All the word vectors are stacked in a word embedding matrix $L \in \mathbb{R}^{d\times \|V\| }$, where $d$ is the dimension of word vector and $V$ is the vocabulary size.

For example, the sentence, ``I lost my phone yesterday, I feel so sad now.'' shown in Figure 1, consists of two clauses. 
The first clause contains the emotion cause while the second clause \added[id=lq]{expresses the emotion of sadness. }
\added[id=lq]{Current methods to emotion cause extraction cannot handle complex sentence structures where the expression of an emotion and its cause are not adjacent. We envision that }
the memory network can \added[id=lq]{better} model the relation between \added[id=lq]{a}
emotion word and \added[id=lq]{its} emotion causes 
in such complex sentence structures. In our approach, we only select the clause with the highest probability to be \replaced[id=lq] {the}{an} emotion cause in each document.

\subsection{Memory Network}
\label{sec:basicMemoryNetwork}

We first present a basic memory network model for emotion cause extraction (shown in Figure 2). Given a clause $c=\{w_1,w_2,...,w_k\}$, and an emotion word, we \added[id=lq]{first obtain the emotion word's representation in}
an embedding space\added[id=lq]{, denoted by $E$}. For the clause, \added[id=lq]{let 
the embedding representations of the words be denoted by }
$e_1,e_2,...,e_k$.
Here, both $e_i$ and $E$ \added[id=lq]{are defined in} $\mathbb{R}^d$. Then, we use the inner product to evaluate the correlation between each word \added[id=lq]{$i$ }in a clause and the emotion word, denoted as $m_i$:
\begin{equation}
m_i = e_i \cdot E.
\end{equation}

We then normalize the value of $m_i$ to $\left[ 0,1 \right]$ using a softmax function, 
denoted by $\alpha_i$ \added[id=lq]{as}:
\begin{equation}
\alpha_i = \frac { \exp { \left(m_i\right) }} {\sum_{j=1}^k \exp \left({m_j} \right)},
\end{equation}
where $k$ is the length of the clause. 
\added[id=lq]{$k$ also serves as} the size of the memory.
\added{Obviously,} $\alpha_i \in \left[ 0,1 \right]$ and $\sum_{i=1}^k {\alpha_i} = 1$. 
\added[id=lq]{$\alpha_i$ can serve as an attention weight}
to measure the importance of each word in our model.

\begin{figure}[htbp]
\label{fig:figure2}
\includegraphics[width=3in]{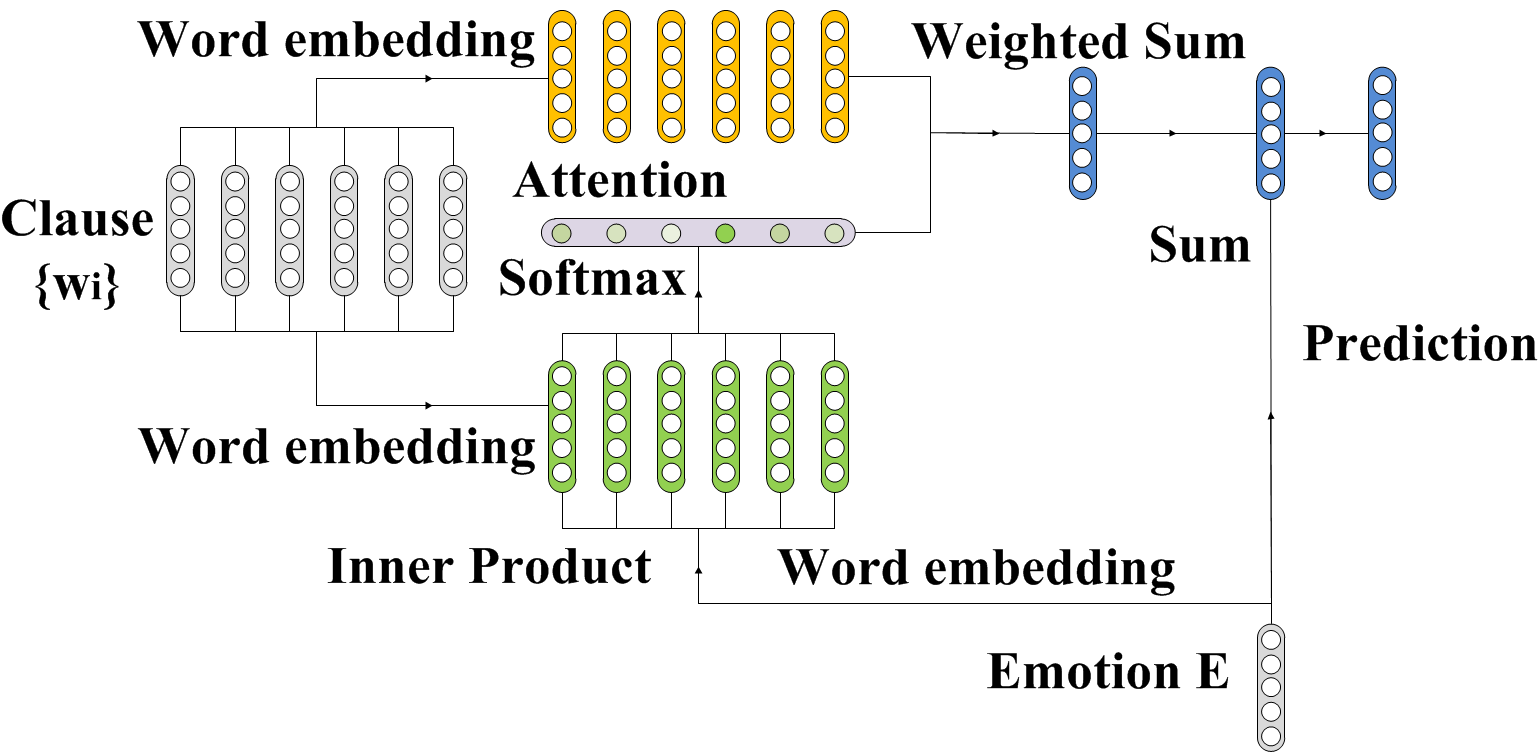}
\caption{A single layer memory network.}
\end{figure}

Then, a sum over the word embedding $e_i$, weighted by the attention vector form the output of the memory network for the prediction of $o$:
\begin{equation}
\ o = \sum_{i=1}^k e_i \cdot \alpha_i + E.
\end{equation}

The final prediction is an output from a softmax function, denoted as $\hat{o}$:
\begin{equation}
\hat{o} = \mbox{softmax} \left( W^T o \right).
\end{equation}

Usually, $W$ is a $d \times d$ weight matrix and $T$ is the transposition. Since the answer in our task is a simple ``yes'' or ``no'', we use a $d \times 1$ matrix for $W$. As the distance between a clause and an emotion words is a very important feature according to \cite{gui2016event}, we simply add this distance into the softmax function as an additional feature in our work. 

\begin{figure}[htbp]
\label{fig:figure3}
\centering
\includegraphics[width=3in]{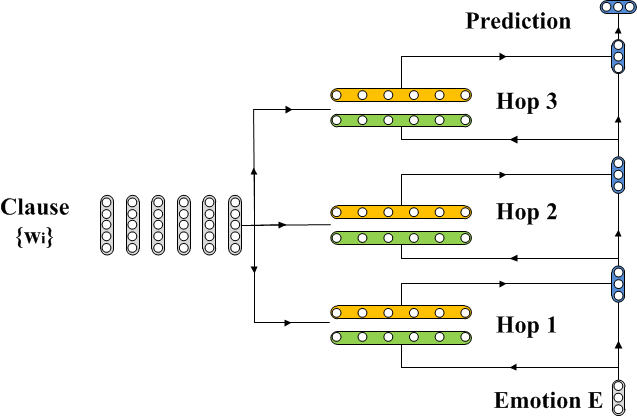}
\caption{Deep memory network with three computational layers (hops).}
\end{figure}

The basic model can be extended to deep architecture consisting of multiple layers to handle $L$ hop operations. The network is stacked as \added[id=lq]{follows:}

\begin{itemize}
\item For hop 1, the query is $E$ and the prediction vector is $o_1$;
\item For hop $i$, the query is the prediction vector of the previous hop and the prediction vector is $o_i$;
\item The output vector is at the top of the network. It is a softmax function on the prediction vector from hop $L$: 
$\hat{o} = \mbox{softmax} \left( W^T o\added[id=lq]{_L} \right)$.
\end{itemize}

The illustration of a deep memory network with three layers is shown in Figure 3. Since \added[id=lq]{a} memory network models the emotion cause 
at a fine-grained level, each word has a corresponding weight to measure 
its importance in this task. Comparing \added[id=lq]{to} previous approaches \added[id=lq]{in} emotion cause extraction which are \added[id=lq]{mostly} based \added[id=lq]{on} manually defined rules or linguistic features, \added[id=lq]{a} memory network is a more principled way to identify the emotion cause from text. However, the basic \added[id=lq]{memory network} model \added[id=lq]{does not capture the sequential information in context which is important in emotion cause extraction.}

\subsection{Convolutional Multiple-Slot Deep Memory Network}
\label{sec:multislots}

It is often the case that the meaning of a word is determined by its context, such as the previous word and the following word. \added[id=lq]{Also, negations and emotion transitions are context sensitive.} However, the memory network described in Section \ref{sec:basicMemoryNetwork} has only one memory slot with size $d \times k$ to represent a clause, where $d$ is the dimension of a word embedding and $k$ is the length of a clause. It means that when the memory network models a clause, it only considers each word separately. 

In order to capture \added[id=lq]{context information for} clauses, we propose a new architecture which contains more memory slot to model the context with a convolutional operation. The basic architecture of Convolutional Multiple-Slot Memory Network (in short: ConvMS-Memnet) is shown in Figure 4.

\begin{figure}[htbp]
\label{fig:figure4}
\centering
\includegraphics[width=3in]{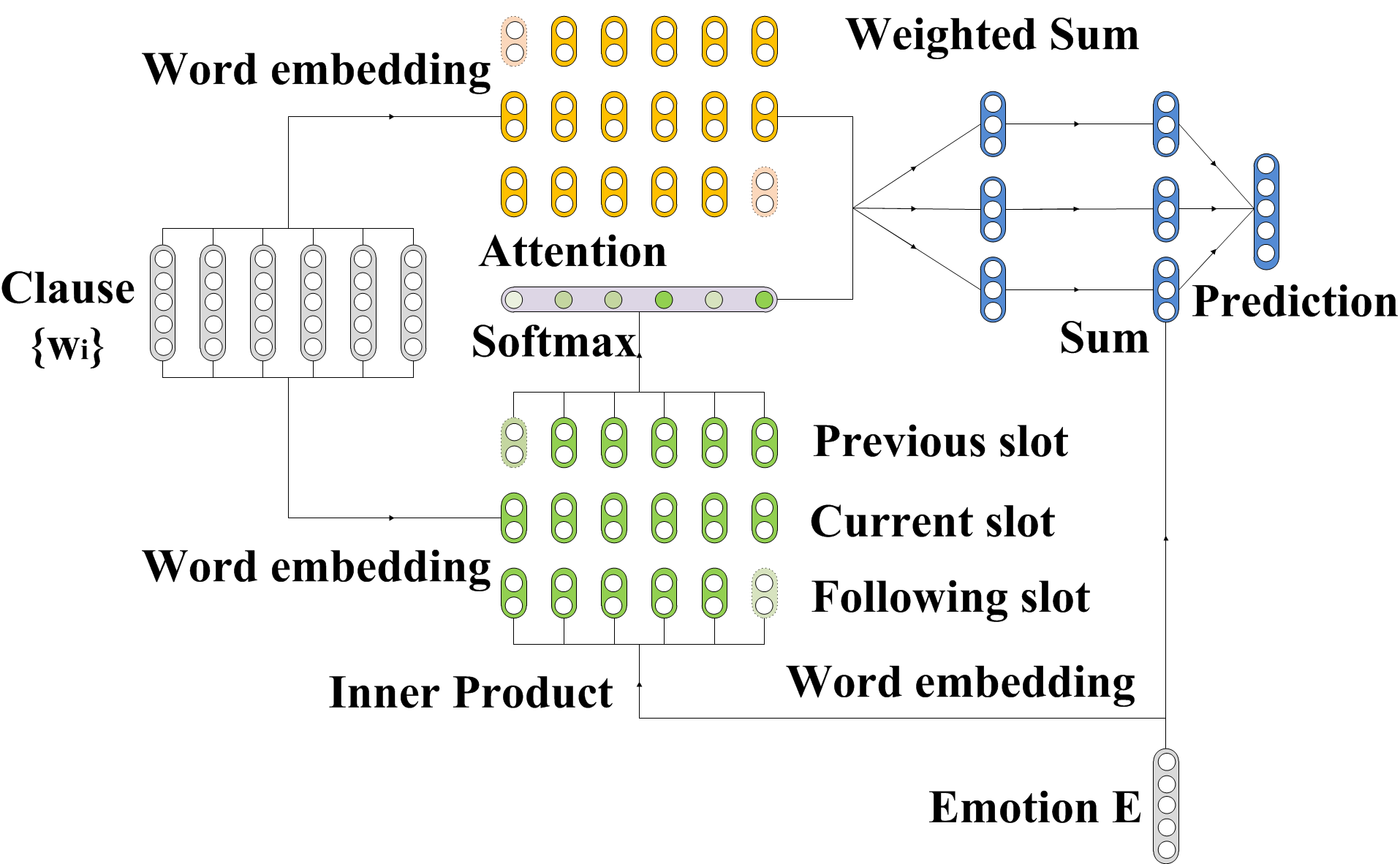}
\caption{A single layer ConvMS-Memnet.}
\end{figure}

Considering the text length is usually short in the dataset used here for emotion cause extraction, we set the size of the convolutional kernel to 3. That is, the weight of word $w_i$ \added[id=lq]{in the $i$-th position} considers both the previous word $w_{i-1}$ and the following word $w_{i+1}$ by a convolutional operation:

\begin{align}
m'_i = \sum^{3}_{j=1} e_{i-2+j} \cdot E
\end{align}

For the first and the last word in a clause, we use zero padding, $w_0 = w_{k+1} = \vec 0$, where $k$ is the length of a clause. Then, the attention \replaced[id=lq]{weight}{signal} for each word position in the clause is \added[id=lq]{now defined as}:
 
\begin{equation}
\alpha'_i = \frac { \exp { \left(m'_i\right) }} {\sum_{j=1}^k \exp \left({m'_j} \right)}
\end{equation}

Note that we obtain the attention for each position rather than each word. It means that the corresponding attention for the $i$-th word in the previous convolutional slot should be $\alpha_{i+1}$. Hence, there are three prediction output vectors, namely, $o_{previous}$, $o_{current}$, $o_{following}$:

\begin{align}
&o_{previous} = \sum_{i=1}^k e_{i-1} \cdot \alpha'_i + E \\
&o_{current} = \sum_{i=1}^k e_i \cdot \alpha'_i + E \\
&o_{following} = \sum_{i=1}^k e_{i+1} \cdot \alpha'_i + E
\end{align}

At last, we concatenate the three vectors as $o = o_{previous} \bigoplus o_{current} \bigoplus o_{following} $ for the prediction by a softmax function:

\begin{equation}
\hat{o} = \mbox{softmax} \left( W_m^T  o  \right) 
\end{equation}

Here, the size of $W_m$ is $ \left( 3 \cdot d \right) \times d $. Since the prediction vector is a concatenation of three outputs. We implement a concatenation operation rather than averaging or other operations because the parameters in different memory slots can be updated \deleted[id=lq]{respectively in this way} by back propagation. The concatenation of three output vectors forms a sequence-level feature which can be used in the training. Such a feature is important especially \added[id=lq]{when the size of annotated training data is small}.

\begin{figure}[htbp]
\label{fig:figure5}
\includegraphics[width=3in]{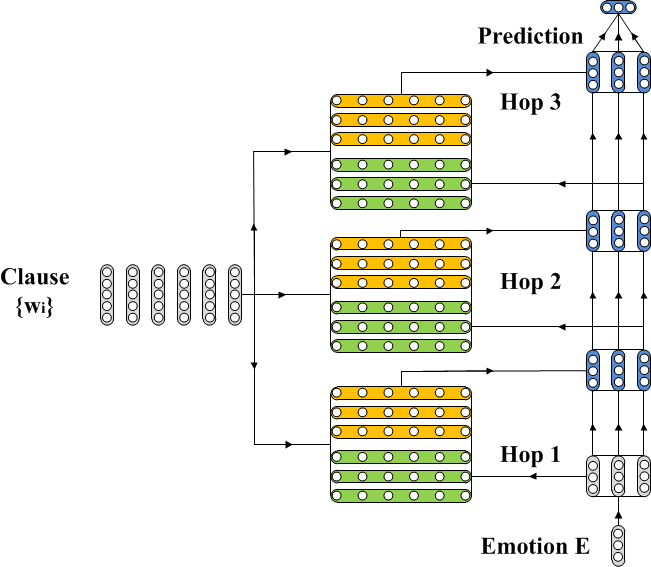}
\caption{ConvMS-Memnet with three computational layers (hops).}
\end{figure}

For deep architecture with multiple layer\deleted[id=lq]{s} training, the network is more \added[id=lq]{complex}
(shown in Figure 5). 

\begin{itemize}
\item For the first layer, the query is an embedding of the emotion word, $E$.
\item In the next layer, there are three input queries since the previous layer has three outputs: $o^1_{previous}$, $o^1_{current}$, $o^1_{following}$. So, for the $j$-th layer ($j \not= 1$), we need to re-define the weight function (5) as:
\end{itemize}

\begin{equation}
m'_i = e_{i-1} \cdot o^{j-1}_{previous} + e_{i} \cdot o^{j-1}_{current} + e_{i+1} \cdot o^{j-1}_{following} 
\end{equation}

\begin{itemize}
\item In the last layer, \added[id=lq]{the concatenation of the three prediction vectors form the final prediction vector to generate the answer.}
\end{itemize}

For model training, we use stochastic gradient descent and back propagation to optimize the loss function. Word embeddings are learned using a skip-gram model. The size of the word embedding is 20 since the vocabulary size in our dataset is small. The dropout is set to 0.4.

\section{Experiments and Evaluation}
\label{sec:section4}

We first presents the experimental settings and then report the results in this section.

\subsection{Experimental Setup and Dataset}

We conduct experiments on a simplified Chinese emotion cause corpus~\cite{gui2016event}\footnote{Available at: http://hlt.hitsz.edu.cn/?page id=694}, 
the only publicly available dataset on this task to the best of our knowledge. The corpus contains 2,105 documents from SINA city news\footnote{http://news.sina.com.cn/society/}. Each document has only one emotion word and one or more emotion causes. The documents \added{are} segmented into clauses manually. The main task is to identify which clause contains the emotion cause.

\begin{table}
\centering
\small
\begin{tabular}{|l|r|}
\hline
{\bf Item} & {\bf Number} \\\hline
Documents & 2,105 \\
Clauses & 11,799 \\
Emotion Causes & 2,167 \\ 
Documents with 1 emotion & 2,046 \\ 
Documents with 2 emotions & 56 \\
Documents with 3 emotions  & 3  \\\hline
\end{tabular} 
\caption{Details of the dataset.}
\end{table}

\added[id=lq]{D}etails of the corpus are shown in Table 1. The metrics we used in evaluation follows \newcite{lee2010text}. It is commonly accepted so that we can compare our results with others. If a proposed emotion cause clause covers the annotated answer, the word sequence is considered correct. The precision, recall, and F-measure are defined by

\begin{align*}
&P=\frac{\sum_{\mbox{correct causes}} 1}{\sum_{\mbox{proposed causes}} 1}, \\
&R=\frac{\sum_{\mbox{correct causes}} 1}{\sum_{\mbox{annotated causes}} 1}, \\
&F=\frac{2 \times P \times R}{P+R}.
\end{align*}

In the experiments, we randomly select 90\% of the dataset as training data and 10\% as testing data. In order to obtain statistically credible results, we evaluate our method and baseline methods 25 times with different train/test splits.

\subsection{Evaluation and Comparison}

We compare with the following baseline methods:

\begin{itemize}
\item \textbf{RB} (Rule based method): The rule based method proposed in~\cite{lee2010text}.
\item \textbf{CB} (Common-sense based method): \added{This is the knowledge based method}
proposed by~\cite{russo2011emocause}. We use the Chinese Emotion Cognition Lexicon~\cite{Xu2013Lexicon} as the common-sense knowledge base. 
The lexicon contains more than 5,000 kinds of emotion stimulation and their corresponding reflection words.
\item \textbf{RB+CB+ML} (Machine learning method trained from rule-based features and facts from a common-sense knowledge base): This methods was previously proposed for emotion cause classification in~\cite{chen2010emotion}. It takes rules and facts in a knowledge base as features for classifier training. We train a SVM using features extracted from the rules defined in~\cite{lee2010text} and the Chinese Emotion Cognition Lexicon~\cite{Xu2013Lexicon}. 
\item \textbf{SVM}: \added{This is a SVM classifier} using the unigram, bigram and trigram features.
It is a baseline previously used in~\cite{li2014text,gui2016event}
\item \textbf{Word2vec}: \added{This is a SVM classifier using word representations learned by Word2vec~\cite{mikolov2013distributed} as features.}
\item \textbf{Multi-kernel}: \added{This is the state-of-the-art method} using the multi-kernel method~\cite{gui2016event} to identify the emotion cause. We use the best performance reported in their paper.
\item \textbf {CNN}: The convolutional neural network for sentence classification~\cite{kim2014con}.
\item \textbf {Memnet}: The deep memory network described in Section \ref{sec:basicMemoryNetwork}. Word embeddings are pre-trained by skip-grams. The number of hops is set to 3.
\item \textbf {ConvMS-Memnet}: The convolutional multiple-slot deep memory network we proposed in Section \ref{sec:multislots}. Word embeddings are pre-trained by skip-grams. The number of hops is 3 in our experiments.
\end{itemize}


\begin{table}
\centering
\small
\begin{tabular}{|l|c|c|c|}
\hline
{\bf Method} & {\bf P} & {\bf R} & {\bf F}\\\hline
RB & \bf0.6747&	0.4287&	0.5243\\
CB  & 0.2672	&\bf0.7130	&0.3887\\
RB+CB & 0.5435&	0.5307&	0.5370\\ 
RB+CB+ML & 0.5921	&0.5307&	0.5597\\ 
SVM &0.4200	&0.4375&	0.4285\\
Word2vec & 0.4301&	0.4233&	0.4136\\ 
{CNN}  & 0.6215&	0.5944&	0.6076\\
Multi-kernel  & 0.6588	&0.6927	&\bf0.6752\\ \hline
{Memnet}  & 0.5922&	0.6354&	0.6131\\
{ConvMS-Memnet} & \bf 0.7076 &	\bf0.6838	&\bf0.6955\\ \hline
\end{tabular} 
\caption{Comparison with existing methods.}
\end{table}


Table 2 shows the \added{evaluation }results.
\added{The rule based} RB gives fairly high precision but with low recall. CB, the common-sense based method, achieves the highest recall.
\added{Yet,} its precision is the worst. \added{RB+CB, the} combination of RB and CB gives higher \added{the F-measure}
But, the improvement of 1.27\% is only marginal compared to RB. 


For machine learning methods, RB+CB+ML uses \added{both} rules and \added{common-sense knowledge} as features to train a machine learning classifier. It achieves F-measure of 0.5597, outperforming RB+CB. 
Both SVM and word2vec are word feature based methods and they have similar performance. For word2vec, even though \added{word representations are}  obtained from the SINA news raw corpus, it still performs worse than SVM trained using n-gram features only. The multi-kernel  method~\cite{gui2016event} is the best performer among the baselines \added{because it considers context information in a structured way.} It models text by its syntactic tree and also considers an emotion lexicon. \added{Their work} shows that the structure information is important for the emotion cause extraction task.

Naively applying \added{the original} deep memory network or convolutional network for emotion cause extraction outperforms all the baselines except the convolutional multi-kernel method. However, using our proposed ConvMS-Memnet architecture, we manage to boost the performance \added{by} 11.54\% in precision, 4.84\% in recall and 8.24\% in F-measure respectively when compared to Memnet. 
The improvement is \added{very} significant with $p$-value less than 0.01 in $t$-test. The ConvMS-Memnet also outperforms the previous best-performing method, multi-kernel, by 3.01\% in F-measure. It shows that by effectively capturing context information, ConvMS-Memnet is able to identify the emotion cause better compared to other methods. 

\subsection{More Insights into the ConvMS-Memnet}

To gain better insights into our proposed ConvMS-Memnet, we conduct further experiments to understand the impact on performance by using: 1) pre-trained or randomly initialized word embedding; 2) multiple hops; 3) attention visualizations; 4) more training epochs.


\subsubsection{Pre-trained Word Embeddings}

\begin{table}
\centering
\small
\begin{tabular}{|l|c|c|c|}
\hline
{\bf Word Embedding} & {\bf P} & {\bf R} & {\bf F}\\\hline
Pre-trained & \bf 0.7076 &	\bf0.6838	&\bf0.6955\\
Randomly initialized  & 0.6786	&0.6608	&0.6696\\ \hline
\end{tabular} 
\caption{Comparison of using pre-trained or randomly initialized word embedding.}
\end{table}

\begin{table}
\centering
\small
\begin{tabular}{|l|c|c|c|}
\hline
{\bf Method} & {\bf P} & {\bf R} & {\bf F}\\\hline
Hop 1 & 0.6597&	0.6444&	0.6520\\
Hop 2  & 0.6877	&0.6718	&0.6796\\
Hop 3 & \bf 0.7076 &	\bf0.6838	&\bf0.6955\\ 
Hop 4 & 0.6882	&0.6722&	0.6801\\ 
Hop 5 &0.6763	&0.6606&	0.6683\\
Hop 6 & 0.6664&	0.6509&	0.6585\\ 
Hop 7  & 0.6483	&0.6333	&0.6407\\ 
Hop 8  & 0.6261&	0.6116&	0.6187\\
Hop 9 &  0.6161 &	0.6109	&0.6089\\ \hline
\end{tabular} 
\caption{Performance with different number of hops in ConvMS-Memnet.}
\end{table}

\begin{table*}[!htbp]
\centering
\small
\begin{tabular}{|l|l|l|c|c|c|c|c|}
\hline
{previous slot} & {\bf current slot} & {following slot} & {\bf Hop 1} & {\bf Hop 2} & {\bf Hop 3} & {\bf Hop 4}  & {\bf Hop 5}\\     \hline
家人/family & 的/'s & 坚持/insisting & \cellcolor{pink}0.1298&	\cellcolor{pink}0.3165&	\cellcolor{pink}0.1781 &\cellcolor{pink}0.2947 &\cellcolor{pink}0.1472\\
的/'s & 坚持/insistence & 更/more  & \cellcolor{pink}0.1706	&\cellcolor{pink}0.2619	&\cellcolor{red}\bf0.7346 &\cellcolor{red}\bf0.6412 &\cellcolor{red}\bf0.8373\\
坚持/insisting &更/more &让/makes  &  \cellcolor{red}\bf0.5090	&\cellcolor{pink}\bf0.3070	&0.0720 &0.0553 &0.0145\\
更/more & 让/makes & 人/people & 0.0327 &	0.0139	&0.0001 &0.0001 &0.0000\\ 
让/makes & 人/people & 感动/touched & \cellcolor{pink}0.1579	&0.0965&	0.0145  &0.0080 &0.0008\\  \hline
\end{tabular} 
\caption{The distribution of attention in different hops.}
\end{table*}

\begin{table}
\centering
\small
\begin{tabular}{|l|c|c|c|}
\hline
{\bf Method} & {\bf P} & {\bf R} & {\bf F}\\\hline
Memnet & 0.5688 &	0.5588	& 0.5635\\
ConvMS-Memnet  & \bf 0.6250 &	\bf0.6140	&\bf0.6195\\ \hline
\end{tabular} 
\caption{Comparison of word level emotion cause extraction.}
\end{table}

\begin{table*}[!htbp]
\centering
\small
\begin{tabular}{|l|c|c|c|c|}
\hline
{\bf Clause} & {\bf 5 Epochs} & {\bf 10 Epochs} & {\bf 15 Epochs} & {\bf 20 Epochs}\\     \hline
45 Days & 0.0018&	0.0002&	0.0000 &0.0000\\
it is ... baby  & 0.3546	&0.6778	&0.5457 &0.3254\\
If the ... back home & \bf 0.7627 &	\bf0.7946	&\bf0.8092 &\bf0.9626\\ 
they ... Spring Festival & 0.2060	&0.0217&	0.0004  &0.0006\\  \hline
\end{tabular} 
\caption{The probability of a clause containing the emotion cause in different iterations in the multiple-slot memory network.}
\end{table*}

In our ConvMS-Memnet, we use pre-trained word embedding as the input. The embedding maps each word into a lower dimensional real-value vector as its representation. Words sharing similar meanings should have similar representations. It enables our model to deal with synonyms more effectively. 

The question is, ``can we train the network without using pre-trained word embeddings?". We initialize word vectors randomly, and use an embedding matrix to update the word vectors in the training of the network simultaneously. Comparison results are shown in Table 3. It can be observed that pre-trained word embedding gives 2.59\% higher F-measure compared to random initialization. This is partly due to the limited size of our training data. Hence using word embedding trained from other much larger corpus gives better results.


\subsubsection{Multiple Hops}
\label{sec:multihops}

It is widely acknowledged that computational models using deep architecture with multiple layers have better ability to learn data representations with multiple levels of abstractions. In this section, we evaluate the power of multiple hops in this task. We set the number of hops from 1 to 9 with 1 standing for the simplest single layer network shown in Figure 4. The more hops are stacked, the more complicated the model is. Results are shown in Table 4.
The single layer network has achieved a competitive performance. With the increasing number of hops, the performance improves. However, when the number of hops is larger than 3, the performance decreases due to overfitting. Since the dataset for this task is small, more parameters will lead to overfitting. As such, we choose 3 hops in our final model since it gives the best performance in our experiments.

\subsubsection{Word-Level Attention Weights}

Essentially, memory network aims to measure the weight of each word in the clause with respect to the emotion word. The question is, will the model really focus on the words which describe the emotion cause? We choose one example to show the attention results in Table 5:

\noindent \textbf{Ex.2} \emph{家人/family 的/'s 坚持/insistence 更/more 让/makes 人/people 感动/touched}

In this example, the cause of the emotion ``touched'' is ``insistence''. We show in Table 5 the distribution of word-level attention weights in different hops of memory network training. We can observe that in the first two hops, the highest attention weights centered on the word ``more". However, from the third hop onwards, the highest attention weight moves to the word sub-sequence centred on the word ``insistence''. This shows that our model is effective in identifying the most important keyword relating to the emotion cause. Also, better results are obtained using deep memory network trained with at least 3 hops. This is consistent with what we observed in Section \ref{sec:multihops}.

In order to evaluate the quality of keywords extracted by memory networks, we define a new metric on the keyword level of emotion cause extraction. The keyword is defined as the word which obtains the highest attention weight in the identified clause. If the keywords extracted by our algorithm is located within the boundary of annotation, it is treated as correct. Thus, we can obtain the precision, recall, and F-measure by comparing the proposed keywords with the correct keywords by:

\begin{align*}
&P=\frac{\sum_{\mbox{correct keywords}} 1}{\sum_{\mbox{proposed keywords}} 1}, \\
&R=\frac{\sum_{\mbox{correct keywords}} 1}{\sum_{\mbox{annotated keywords}} 1}, \\
&F=\frac{2 \times P \times R}{P+R}.
\end{align*}

Since the reference methods do not focus on the keywords level, we only compare the performance of Memnet and ConvMS-Memnet in Table 6. It can be observed that our proposed ConvMS-Memnet outperforms Memnet by 5.6\% in F-measure. It shows that by capturing context features, ConvMS-Memnet is able to identify the word level emotion cause better compare to Memnet.


\subsubsection{Training Epochs}
\label{sec:trainIterations}

In our model, the training epochs are set to 20. In this section, we examine the testing error using a case study. Due to the page length limit, we only choose one example from the corpus. The text below has four clauses:

\noindent \textbf{Ex.3} \emph{45天，对于失去儿子的他们是多么的漫长，宝贝回家了，这个春节是多么幸福。}

\noindent \textbf{Ex.3} \emph{45 days, it is long time for the parents who lost their baby. If the baby comes back home, they would become so happy in this Spring Festival.}

In this example, the cause of emotion ``happy'' is described in the third clause. 

We show in Table 7 the probability of each clause containing an emotion cause in different training epochs. It is interesting to see that our model is able to detect the correct clause with only 5 epochs. With the increasing number of training epochs, the probability associated with the correct clause increases further while the probabilities of incorrect clauses decrease generally.


\subsection{Limitations}
\label{sec:limitations}

We have shown in Section \ref{sec:trainIterations} a simple example consisting of only four clauses from which our model can identify the clause containing the emotion cause correctly. We notice that for some complex text passages which contain long distance dependency relations, negations or emotion transitions, our model may have a difficulty in detecting the correct clause containing the emotion causes. 
It is a challenging task to properly model the discourse relations among clauses. In the future, we will explore different network architecture with consideration of various discourse relations possibly through transfer learning of larger annotated data available for other tasks. 

Another shortcoming of our model is that, the answer generated from our model is simply ``yes'' or ``no''. The main reason is that the size of the annotated corpus is too small to train a model which can output natural language answers in full sentences. Ideally, we would like to develop a model which can directly give the cause of an emotion expressed in text. However, since the manual annotation of data is too expensive for this task, we need to explore feasible ways to automatically collect annotate data for emotion cause detection. We also need to study effective evaluation mechanisms for such QA systems.

\section{Conclusions}
\label{sec:section5}

In this \added[id=lq]{work,}
we \added[id=lq]{treat}
emotion cause extraction as a QA task and propose a new model based on deep memory networks for identifying 
\added[id=lq]{the emotion causes}
for an emotion expressed in text. 
\added[id=lq]{The key property of this approach is the use of context information in the learning process which is ignored in the original memory network.}
Our new \added[id=lq]{memory network} architecture is able \added[id=lq]{to store}
context in different memory slots to capture 
context information \added[id=lq]{in proper sequence} by convolutional operation. Our model achieves the state-of-the-art performance on a dataset for emotion cause detection when compared to a number of competitive baselines. In the future, we will explore effective ways \added[id=lq]{to model}
discourse relations among clauses and develop a QA system which can directly output the cause of emotions as answers.


\section*{Acknowledgments}

This work was supported by the National Natural Science Foundation of China 61370165, U1636103, 61632011, 61528302, National 863 Program of China 2015AA015405, Shenzhen Foundational Research Funding JCYJ20150625142543470, JCYJ20170307150024907 and Guangdong Provincial Engineering Technology Research Center for Data Science 2016KF09.

\bibliography{emnlp2017}
\bibliographystyle{emnlp_natbib}
\clearpage\end{CJK*}
\end{document}